\pdfoutput=1
\documentclass[11pt]{article}

\usepackage{acl}

\usepackage{times}
\usepackage{latexsym}

\usepackage[T1]{fontenc}

\usepackage[utf8]{inputenc}

\usepackage{microtype}

\usepackage{todonotes}
\usepackage{booktabs}
\usepackage{multirow}
\usepackage{cleveref}
\usepackage[british]{babel}  % rule britannia
\usepackage[nolist,nohyperlinks]{acronym}
\usepackage{caption}
\usepackage{subcaption}
\usepackage{soul}

\title{Exploring Diversity in Back Translation for Low-Resource Machine Translation}

\author{Laurie Burchell \and Alexandra Birch \and Kenneth Heafield \\
    Institute for Language, Cognition, and Computation \\
    School of Informatics, University of Edinburgh \\
    10 Crichton Street, Edinburgh, EH8 9AB, UK \\
    \texttt{\{laurie.burchell,a.birch,kenneth.heafield\}@ed.ac.uk} \\
}

\begin{acronym}
  \acro{BT}{back translation}
  \acro{MAP}{maximum a-posteriori}
  \acro{NMT}{neural machine translation}
\end{acronym}
  
\begin{document}

\maketitle

\begin{abstract}

Back translation is one of the most widely used methods for improving the performance of neural machine translation systems. Recent research has sought to enhance the effectiveness of this method by increasing the `diversity' of the generated translations. We argue that the definitions and metrics used to quantify `diversity' in previous work have been insufficient. This work puts forward a more nuanced framework for understanding diversity in training data, splitting it into lexical diversity and syntactic diversity. We present novel metrics for measuring these different aspects of diversity and carry out empirical analysis into the effect of these types of diversity on final neural machine translation model performance for low-resource English$\leftrightarrow$Turkish and mid-resource English$\leftrightarrow$Icelandic. Our findings show that generating back translation using nucleus sampling results in higher final model performance, and that this method of generation has high levels of both lexical and syntactic diversity. We also find evidence that lexical diversity is more important than syntactic for back translation performance.

\end{abstract}

\section{Introduction}

The data augmentation technique of \ac{BT} is used in nearly every current \ac{NMT} system to reach optimal performance \citep[][\textit{inter alia}]{edunov-etal-2020-evaluation,barrault-etal-2020-findings,akhbardeh-etal-2021-findings}. It involves creating a pseudo-parallel dataset by translating target-side monolingual data into the source language using a secondary \ac{NMT} system \citep{sennrich-etal-2016-improving}. In this way, it enables the incorporation of monolingual data into the \ac{NMT} system. Whilst adding data in this way helps nearly all language pairs, it is particularly important for low-resource \ac{NMT} where parallel data is scarce by definition.  

% (Start with motivation - vast majority of NMT systems use BT and it really helps - very widely use. Want to say BT is ubiquitous and really important - used in NLP in general)

% Data augmentation approaches such as BT allow us to use monolingual data to improve low-resource machine translation/ machine translation when parallel data is scarce

% The vast majority of language pairs require \ac{BT} to reach optimal performance \ac{NMT} 

% % Training a high-quality \ac{NMT} system relies on large amounts of high-quality parallel data. This allows the system to learn a model which will generalise well and so achieve higher task performance. However, most language pairs do not have such resources easily available. For this reason, \ac{NMT} systems for low-resource languages generally rely on data-augmentation techniques \citep{haddow-etal-2021-survey} which involve creating synthetic parallel data to improve performance. 

% The most common data augmentation technique currently is \ac{BT}, which involves creating a pseudo-parallel dataset by translating target-side monolingual data into the source language using a secondary \ac{NMT} system \citep{sennrich-etal-2016-improving}. 

Because of its ubiquity, there has been extensive research into how to improve \ac{BT} \citep{burlot-yvon-2018-using, hoang-etal-2018-iterative, fadaee-monz-2018-back, caswell-etal-2019-tagged}, especially in ways which increase the `diversity' of the back-translated dataset \citep{edunov-etal-2018-understanding, soto-etal-2020-selecting}. Previous work \citep{gimpel-etal-2013-systematic, ott-etal-2018-analyzing, vanmassenhove-etal-2019-lost} has found that machine translations lack the diversity of human productions. This is because most translation systems use some form of \ac{MAP} estimation, meaning that they will always favour the most probable output. \citet{edunov-etal-2018-understanding} and \citet{soto-etal-2020-selecting} argue that this makes standard \ac{BT} data worse training data since it lacks `richness' or diversity. 
 
Despite the focus on increasing diversity in \ac{BT}, what `diversity' actually means in the context of \ac{NMT} training data is ill-defined. In fact,  \citet{tevet-berant-2021-evaluating} point out that there is no standard metric for measuring diversity.  Most previous work uses the BLEU score between candidate sentences or another n-gram based metric to estimate similarity \citep{zhu-etal-2016-texygen, hu2019parabank, he-etal-2018-sequence, shen-etal-2019-mixture,shu-etal-2019-generating, Holtzman2020The, thompson-post-2020-paraphrase}. However, such metrics mostly measure changes in the vocabulary or spelling. Because of this, they are likely to be less sensitive to other kinds of variety such as changes in structure.

We argue that quantifying `diversity' using n-gram based metrics alone is insufficient. Instead, we split diversity into two aspects: variety in the word choice and spelling, and variety in structure. We call these aspects \textit{lexical diversity} and \textit{syntactic diversity} respectively. Here, we follow recent work in natural language generation and particularly paraphrasing \citep[e.g.][]{iyyer-etal-2018-adversarial,krishna-etal-2020-reformulating,goyal-durrett-2020-neural,huang-chang-2021-generating, hosking-lapata-2021-factorising} which explicitly models the meaning and form of the input separately. Of course, there are likely more kinds of diversity than this, but this distinction provides a common-sense framework to extend our understanding of the concept. To our knowledge, no other previous work in data augmentation has attempted to isolate and automatically measure syntactic and lexical diversity.

Building from our definition, we introduce novel metrics aimed at measuring lexical and syntactic diversity separately. We then carry out an empirical study into what effect training data with these two kinds of diversity has on final \ac{NMT} performance in the context of low-resource machine translation. We do this by creating \ac{BT} datasets using different generation methods and measuring their diversity. We then evaluate what impact different aspects of diversity have on final model performance. We find that a high level of diversity is beneficial for final \ac{NMT} performance, though lexical diversity seems more important than syntactic diversity. Importantly though there are limits to both; the data should not be so `diverse' that it affects the adequacy of the parallel data.

We summarise our contributions as follows:
\begin{itemize}
    \item We put forward a more nuanced definition of `diversity' in \ac{NMT} training data, splitting it into \textit{lexical diversity} and \textit{syntactic diversity}. We present two novel metrics for measuring these different aspects of diversity.
    \item We carry out empirical analysis into the effect of these types of diversity on final \ac{NMT} model performance for low-resource English$\leftrightarrow$Turkish and mid-resource English$\leftrightarrow$Icelandic.
    \item We find that nucleus sampling is the highest-performing method of generating \ac{BT}, and it combines both lexical and syntactic diversity.
    \item We make our code publicly available.\footnote{\url{github.com/laurieburchell/exploring-diversity-bt}}
\end{itemize}

\section{Methods}
\label{sec:methods}

We explain each method we use for creating diverse \ac{BT} datasets in \Cref{sec:generating}, then discuss our metrics for diversity in \Cref{sec:metrics}.

\subsection{Generating diverse back translation}
\label{sec:generating}

We use four methods to generate diverse \ac{BT} datasets: beam search, pure sampling, nucleus sampling, and syntax-group fine-tuning. The first three were chosen because they are in common use and so more relevant for future work. The last, syntax-group fine-tuning, aims to increase syntactic diversity specifically and so allows us to separate its effect on final \ac{NMT} performance from lexical diversity. For each method, we create a diverse \ac{BT} dataset by generating three candidate translations for each input sentence. This allows us to measure diversity whilst keeping the `meaning' of the sentence as similar as possible. In this way, we measure inter-sentence diversity as a proxy for the diversity of the dataset as a whole. We discuss our datasets in detail in \Cref{sec:data}.

\paragraph{Beam search} 
Beam search is the most common search algorithm used to decode in \ac{NMT} systems. Whilst it is generally successful in finding a high-probability output, the translations it produces tend to lack diversity since it will always default to the most likely alternative in the case of ambiguity \citep{ott-etal-2018-analyzing}. We use beam search to generate three datasets for each language pair, using a beam size of five and no length penalty: 
\begin{itemize}
    \item \textit{base}: three million input sentences used to generate one output per input (\ac{BT} dataset length: three million)
    \item \textit{beam}: three million input sentences used to generate three outputs per input (\ac{BT} dataset length: nine million)
    \item \textit{base-big}: nine million input sentences used to generate one output per output (\ac{BT} dataset length: nine million)
\end{itemize}

\paragraph{Pure sampling}
An alternative to beam search is sampling from the model distribution. At each decoding step, we sample from the learned distribution without restriction to generate output. This method means we are likely to generate a much wider range of tokens than restricting our choice to those which are most likely (as in beam search). However, it also means that the generated text is less likely to be adequate (have the same meaning as the input) as the output space does not necessary restrict itself to choices which best reflect the meaning of the input. In other words, the output may be diverse, but it may not be the kind of diversity that we want for \ac{NMT} training data.  

We create one dataset per language pair (\textit{sampling}) by generating three candidate translations for each of the three million monolingual input sentences. This results in nine-million line \ac{BT} dataset. We set our beam size to five when generating.

\paragraph{Nucleus sampling}
Nucleus or top-p sampling is another sampling-based method, introduced by \citet{Holtzman2020The}. Unlike pure sampling, which samples from the entire distribution, top-p sampling only samples from the highest probability tokens whose cumulative probability mass exceeds the pre-chosen threshold \textit{p}. The intuition is that when only a small number of tokens are likely, we want to limit our sampling space to those. However, when there are many likely hypotheses, we want to widen the number of tokens we might sample from. We chose this method in the hope it represents a middle ground between high-probability but repetitive beam search generations, and more diverse but potentially low-adequacy pure sampling generation. We create one dataset per language pair (\textit{nucleus}) by generating three hypothesis translations for each of the three million monolingual input sentences. Each dataset is therefore nine million lines long. We set the beam size to five and \textit{p} to 0.95.

\paragraph{Syntax-group fine-tuning}
For our analysis in this paper, we want to generate diverse \ac{BT} in a way which focuses on syntactic diversity over lexical diversity, so that we can separate out its effect on final \ac{NMT} performance. We therefore take a fine-tuning approach for our final generation method. To do this, we generate the dependency parse of each sentence in the English side of the parallel data for each language pair using the Stanford neural network dependency parser \citep{chen-manning-2014-fast}. We then label each pair of parallel sentences in the training data according to the first split in the corresponding syntactic parse tree. We then create three fine-tuning training datasets out of the three biggest syntactic groups.\footnote{For English--Turkish, we combine the third and fourth largest syntactic groups to create the third fine-tuning dataset, as the third-largest syntactic group alone was not large enough for successful fine-tuning.} Finally, we take \ac{NMT} models trained on parallel data alone and restart training on each syntactic-group dataset, resulting in three \ac{NMT} systems which are fine-tuned to produce a particular syntactic structure. We are only able to create models this way which translate into English, as good syntactic parsers are not available for the other languages in our study.

\begin{figure}
    \centering
    \includegraphics[width=0.5\textwidth]{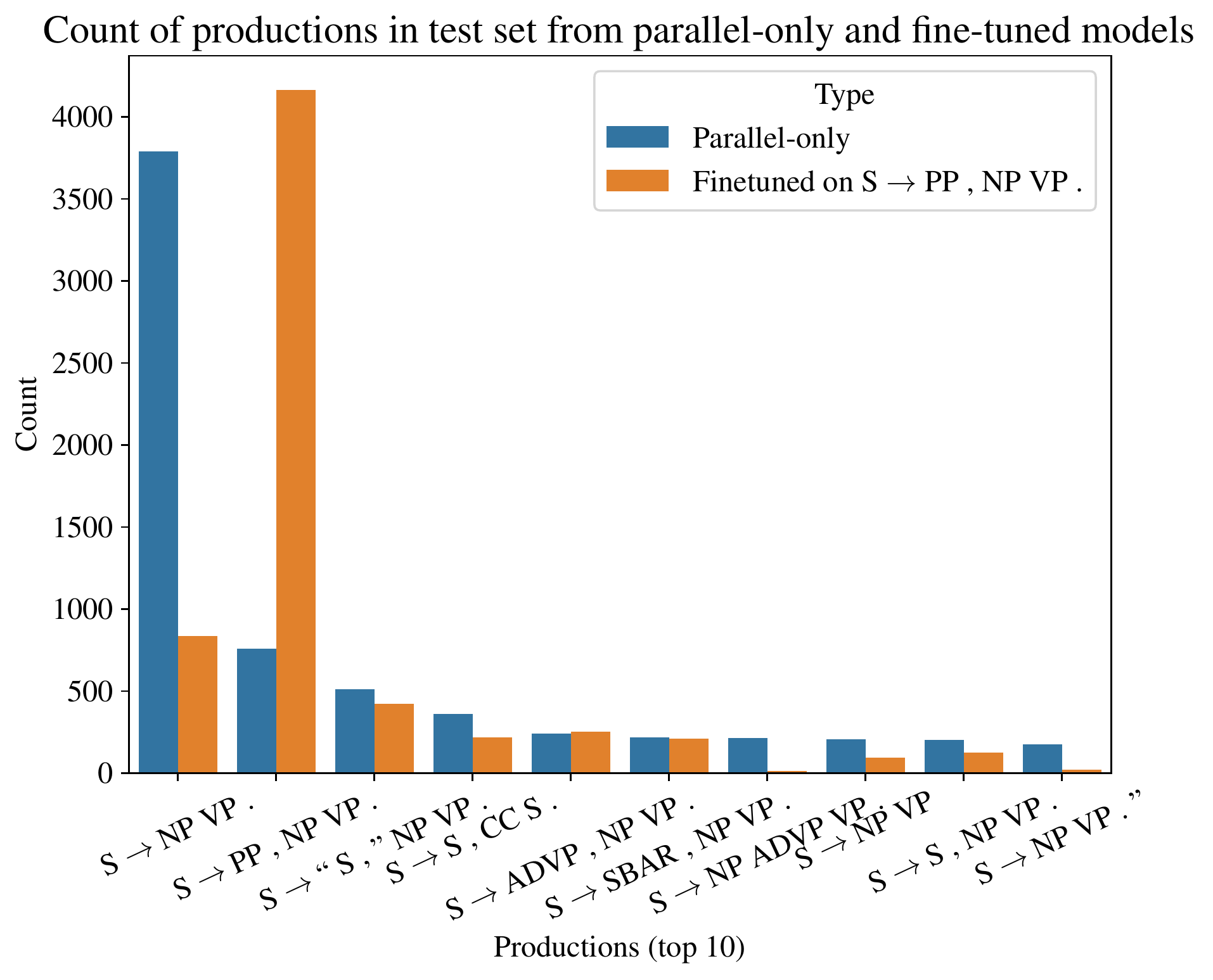}
    \caption{The count of the top-ten syntactic groups produced by the parallel-only Turkish$\rightarrow$English NMT model compared to the number of those productions produced by a Turkish$\rightarrow$English NMT model fine-tuned on the second-most common syntactic group (\texttt{S -> PP NP VP .}). The fine-tuned model produces more examples of the required syntactic group. Input data is the combined WMT test sets.}
    \label{fig:finetune}
\end{figure}

To verify this method works as expected, we translated the test set for each language pair with the model trained on parallel data only. We then translated the same test set with each fine-tuned model and checked it was producing more of the required syntactic group. We did indeed find that fine-tuning resulted in more candidate sentences from the required group. \Cref{fig:finetune} gives an example of the different pattern of productions between the parallel-only model and a model fine-tuned on a particular syntactic group (\texttt{S -> PP NP VP .})

\subsection{Diversity metrics}
\label{sec:metrics}

We use three primary metrics to measure lexical and syntactic diversity: i-BLEU, i-chrF, and tree kernel difference. As mentioned in \Cref{sec:generating}, we generate three output sentences for each input to our \ac{BT} systems and measure inter-sentence diversity as a proxy for the diversity produced by the system.  Due to compute time, we calculate all inter-sentence metrics over a sample of 30,000 sentence groups rather than the whole \ac{BT} dataset.

% \citet{edunov-etal-2018-understanding} found that generating \ac{BT} with sampling improved final \ac{NMT} model performance compared to beam search. They argue that sampling better approximates the data distribution \citep{ott-etal-2018-analyzing}, and thus is ``more likely to provide a richer training signal than argmax sequences''. We extend this analysis by investigating the characteristics of this richer training data with respect to its diversity. Following recent work on paraphrasing \citep{iyyer-etal-2018-adversarial, hosking-lapata-2021-factorising, huang-chang-2021-generating}, we split diversity into two aspects: lexical (variation in form) and syntactic (variation in structure). There are likely more kinds of diversity, but this definition provides a common-sense framework to extend our understanding of the concept.

% We also calculate three summary statistics over the whole generated dataset: mean sentence length, mean word length, and total vocabulary size. We give a brief explanation of each of these metrics and how they are calculated below. We generate three output sentences for each input to our \ac{BT} systems and measure inter-sentence diversity as a proxy for the diversity produced by the system. 

\paragraph{i-BLEU}
Following previous work, we calculate the BLEU score between all sentence pairs generated from the same input \citep{papineni-etal-2002-bleu}, take the mean and then subtract it from one to give inter-sentence or i-BLEU \citep{zhu-etal-2016-texygen}. We believe that lexical diversity as we define it is the main driver of this metric, since BLEU scores are calculated based on n-gram overlap and so the biggest changes to the score will result from changes to the words used (though changes in ordering of words and their morphology will also have an effect). The higher the i-BLEU score, the higher the diversity of output.

\paragraph{i-chrF}
Building from i-BLEU, we introduce i-chrF, which is generated in the same way as i-BLEU but using chrF \citep{popovic-2015-chrf}. Since chrF is also based on n-gram overlap, we believe it will also mostly measure lexical diversity. However, i-chrF is based on character rather than word overlap, and so should be less affected by morphological changes to the form of words than i-BLEU. We calculate both chrF and BLEU scores using the sacreBLEU toolkit \citep{post-2018-call}.

\paragraph{Tree kernel difference}
We propose a novel metric which focuses on syntactic diversity: mean tree kernel difference. To calculate it, we first generate the dependency parse of each candidate sentence using the Stanford neural network dependency parser \citep{chen-manning-2014-fast}. We replace all terminals with a dummy token to minimise the effect of lexical differences, then we calculate the tree kernel for each pair of parses using code from \citet{conklin-etal-2021-meta}, which is in turn based on \citet{moschitti-2006-making}. Finally, we calculate the mean across all pairs to give the mean tree kernel difference for each set of generated sentences.

 We are only able to calculate the tree kernel metric for the English datasets due to the lack of reliable parsers in Turkish and Icelandic, though this method could extend to any language with a reasonable parser available. The higher the score, the higher the diversity of the output.

\paragraph{Summary statistics}
We calculate mean word length, mean sentence length, and vocabulary size over the entire generated dataset as summary statistics. We use the definition of `word' as understood by the bash \texttt{wc} command to calculate all metrics, since we are only interested in a rough measure to check for degenerate results.

\section{Experiments}

Having discussed the methods by which we generate diverse \ac{BT} datasets and the metrics with which we measure the diversity in these datasets, we now outline our experimental set up for testing the effect of training data diversity on final \ac{NMT} model performance.

\subsection{Data and preprocessing}
\label{sec:data}

We carry out our experiments on two language pairs: low-resource Turkish--English and mid-resource Icelandic--English. These languages are sufficiently low-resource that augmenting the training data will likely be beneficial, but well-resourced enough that we can still train a reasonable back-translation model on the available parallel data alone.

\paragraph{Data provenance}
The Turkish--English parallel data is from the WMT 2018 news translation task \citep{bojar-etal-2018-findings}. The training data is from the SETIMES dataset, a parallel dataset of news articles in Balkan languages \citep{tiedemann-2012-parallel}. We use the development set from WMT 2016 and the test sets from WMT 2016--18.

The Icelandic--English parallel data is from the WMT 2021 news translation task \citep{akhbardeh-etal-2021-findings}. There are four sources of training data: ParIce \citep{barkarson-steingrimsson-2019-compiling}, filtered as described in \citet{jonsson2020experimenting}; Paracrawl \citep{banon-etal-2020-paracrawl}; WikiMatrix \citep{schwenk-etal-2021-wikimatrix}; and WikiTitles\footnote{\url{data.statmt.org/wikititles/v3}}. We use the development and test sets provided for WMT 2021.

The English monolingual data is made up of news crawl data from 2016 to 2020, version 16 of news-commentary crawl,\footnote{\url{data.statmt.org/news-commentary/v16}} and crawled news discussions from 2012 to 2019.\footnote{\url{data.statmt.org/news-discussions/en}} The Turkish monolingual data is news crawl data from 2016 to 2020.\footnote{\url{data.statmt.org/news-crawl}} The Icelandic monolingual data is made up of news crawl data from 2020, and part one of the Icelandic Gigaword dataset \citep{steingrimsson-etal-2018-risamalheild}.

\paragraph{Data cleaning} 
Our cleaning scripts are adapted from those provided by the Bergamot project.\footnote{\url{github.com/browsermt/students/tree/master/train-student}} The full data preparation procedure is provided in the repo accompanying this paper. After cleaning, the Turkish--English parallel dataset contains 202 thousand lines and the Icelandic--English parallel dataset contains 3.97 million lines. The English, Icelandic, and Turkish cleaned monolingual datasets contain 487 million, 39.9 million, and 26.1 million lines respectively. We select 9 million lines of each monolingual dataset for \ac{BT} at random since all the monolingual datasets are the same domain as the test sets.

\paragraph{Text pre-processing} 
We learn a joint BPE model with SentencePiece using the concatenated training data for each language pair \citep{kudo-richardson-2018-sentencepiece}. We set vocabulary size to 16,000 and character coverage to 1.0. All other settings are default. We apply this model to the training, development, and test data. We remove the BPE segmentation before calculating any metrics.

\begin{figure}[t!]
    \centering
    \small
    \includegraphics[width=\linewidth]{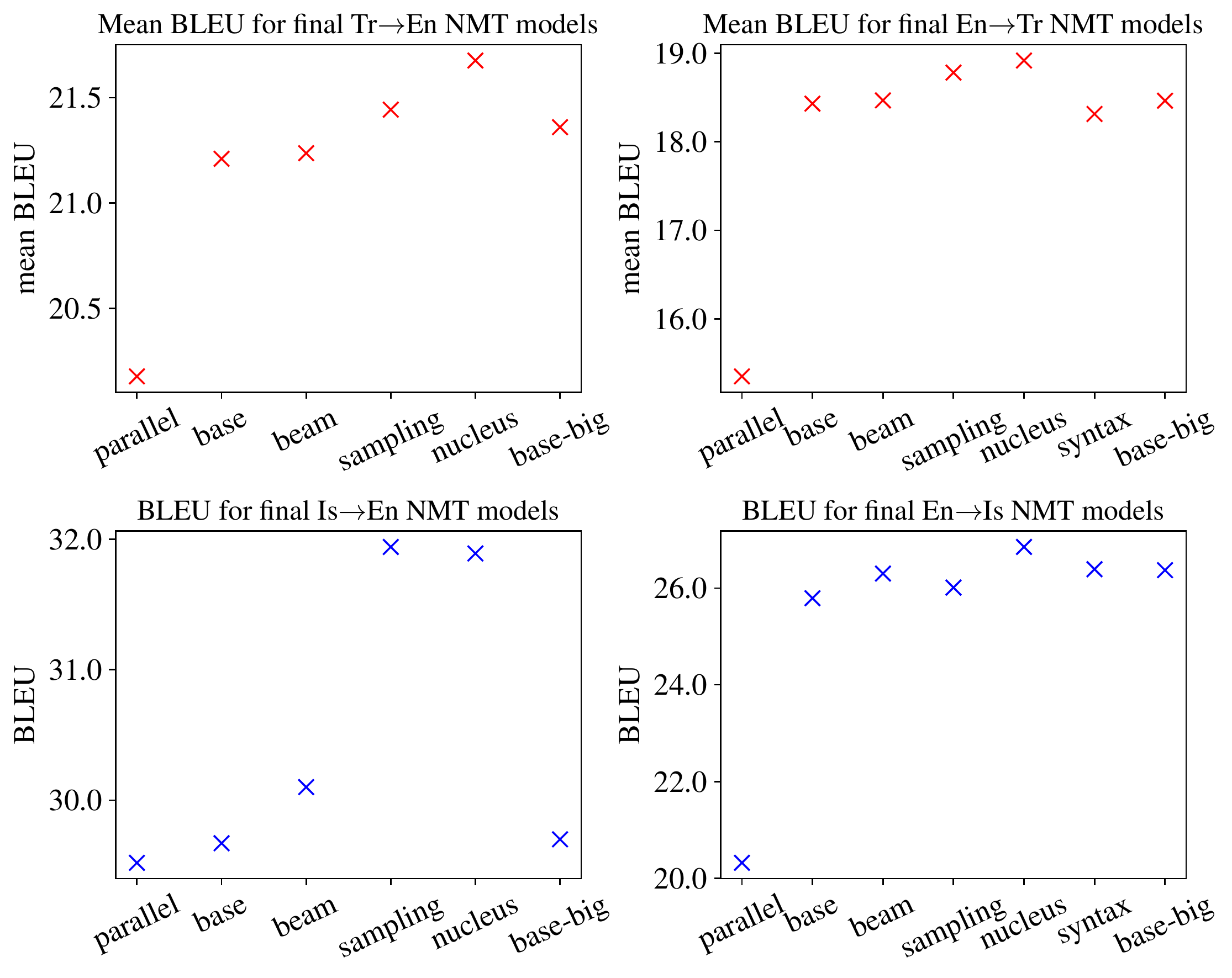}
    \caption{Mean BLEU score on WMT test sets for English$\leftrightarrow$Turkish and English$\leftrightarrow$Icelandic models trained on different BT datasets. For English$\leftrightarrow$Turkish, we give the mean score on WMT 16, WMT 17, and WMT 18 test sets. For English$\leftrightarrow$Icelandic, we give the score on the WMT 21 test set.}
    \label{fig:perf_bleu}
\end{figure}

\subsection{Model training}
\label{sec:method}

\paragraph{Model architecture and infrastructure} All \ac{NMT} models in this paper are transformer models \citep{vaswani-etal-2017-attention}. We give full details about hyper-parameters and infrastructure in \Cref{sec:model_desc}.

\paragraph{Parallel-only models for back translation}
For each language pair and in both directions, we train an NMT model on the cleaned parallel data alone using the relevant hyper-parameter settings in \Cref{tab:hyp}. We measure the performance of these models by calculating the BLEU score \citep{papineni-etal-2002-bleu} using the sacreBLEU toolkit \citep{post-2018-call}\footnote{\texttt{BLEU|nrefs:1|case:mixed|eff:no|\\tok:13a|smooth:exp|version:2.0.0}} and by evaluating the translations with COMET using the \texttt{wmt20-comet-da} model \citep{rei-etal-2020-comet}.

\paragraph{Generating back translation}
For each language pair and in each direction, we use the trained parallel-only models to generate back translation datasets as described in \Cref{sec:generating}. We translate the same three million sentences of monolingual data each time for consistency, translating an additional six million lines of monolingual data for the \textit{base-big} dataset.

\begin{figure}[t!]
    \centering
    \small
    \includegraphics[width=\linewidth]{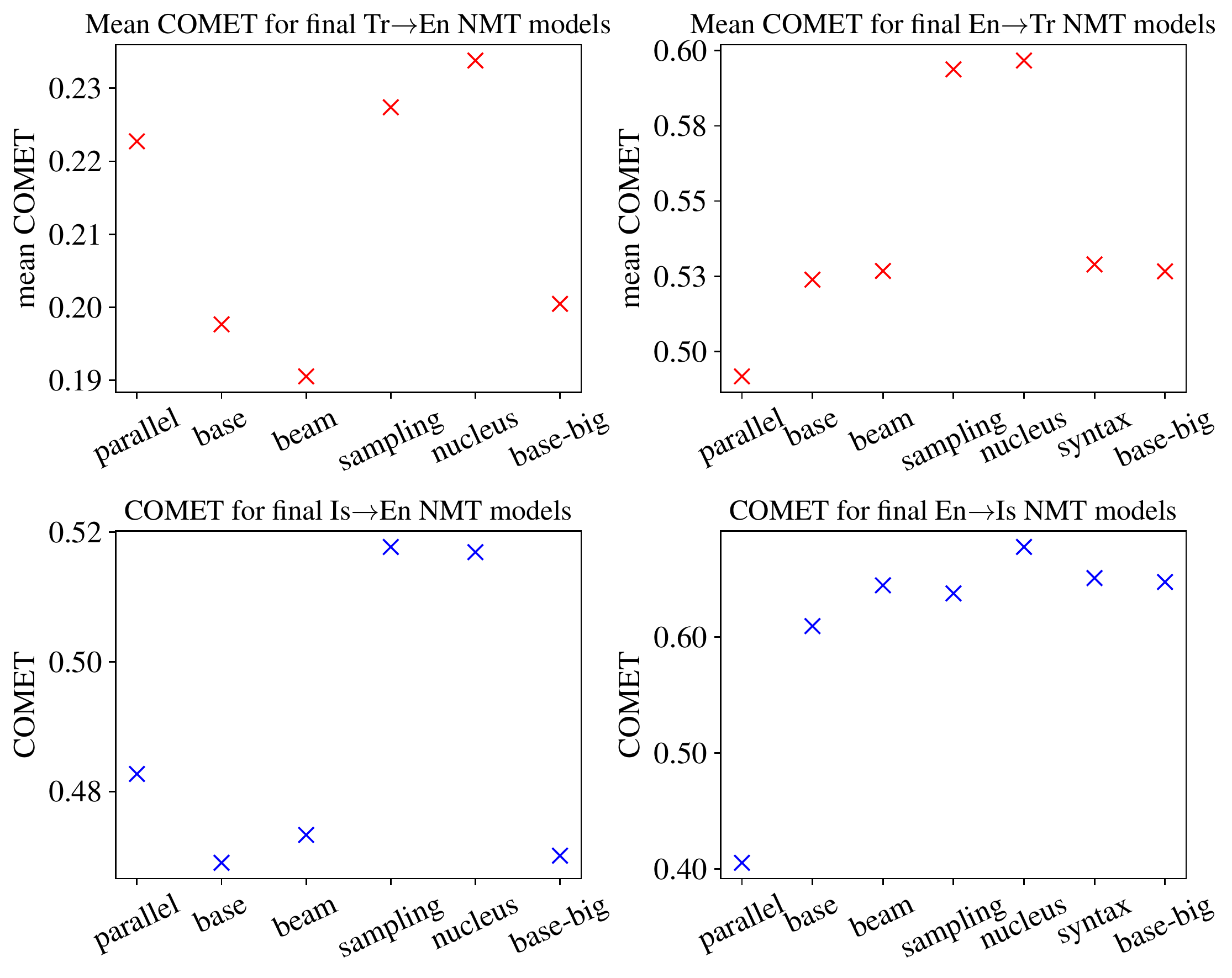}
    \caption{Mean COMET score on WMT test sets for English$\leftrightarrow$Turkish and English$\leftrightarrow$Icelandic models trained on different BT datasets. For English$\leftrightarrow$Turkish, we give the mean score on WMT 16, WMT 17, and WMT 18 test sets. For English$\leftrightarrow$Icelandic, we give the score on the WMT 21 test set.}
    \label{fig:perf_comet}
\end{figure}

\paragraph{Training final models}
We train final models for each language direction on the concatenation of the parallel data and each back-translation dataset (back-translation on the source side, original monolingual data as target). We measure the final performance of these models using BLEU and COMET as before.

\section{Results and Analysis}

\subsection{Final model performance}

% \begin{figure}[htbp]
%     \centering
%     \small
%     \includegraphics[width=\linewidth]{perf_bleu.pdf}
%     \caption{Mean BLEU score on WMT test sets for English$\leftrightarrow$Turkish and English$\leftrightarrow$Icelandic models trained on different BT datasets. For English$\leftrightarrow$Turkish, we give the mean score on WMT16, WMT17, and WMT18 test sets. For English$\leftrightarrow$Icelandic, we give the score on the WMT21 test set.}
%     \label{fig:perf_bleu}
% \end{figure}

% \begin{figure}[htbp]
%     \centering
%     \small
%     \includegraphics[width=\linewidth]{perf_comet.pdf}
%     \caption{Mean COMET score on WMT test sets for English$\leftrightarrow$Turkish and English$\leftrightarrow$Icelandic models trained on different BT datasets. For English$\leftrightarrow$Turkish, we give the mean score on WMT16, WMT17, and WMT18 test sets. For English$\leftrightarrow$Icelandic, we give the score on the WMT21 test set.}
%     \label{fig:perf_comet}
% \end{figure}

\Cref{fig:perf_bleu,fig:perf_comet} show the mean BLEU and COMET scores achieved by the final models trained on the concatenation of the parallel data and the different \ac{BT} datasets. In most cases, adding any \ac{BT} data to the training data results in some improvement over the parallel-only baseline for both scores. However, augmenting the training data with \ac{BT} produced with nucleus sampling nearly always results in the strongest performance, with mean gains of 2.88 BLEU or 0.078 COMET. This compares to mean gains of 2.24 BLEU or 0.026 COMET when using the baseline \ac{BT} dataset of three million lines translated with beam search. Pure sampling tends to perform similarly but not quite as well as nucleus sampling. Based on this result, we suggest that future work generate \ac{BT} with nucleus sampling rather than pure sampling.

\begin{table}[t!]
\small
    \centering
    \begin{tabular}{rcccc}
        Dataset & \textit{base-big} & \textit{beam} & \textit{sampling} & \textit{nucleus} \\
        \midrule
        Sent. len. & 12.23 & 12.21 & 13.11 & 12.74 \\
        Word. len.  & 8.19 & 8.17 & 8.37 & 8.28 \\
        Vocab & 1.6M & 1.0M & 5.6M & 3.4M \\
        \midrule
        i-BLEU & - & 38.11 & 86.69 & 83.27 \\
        i-chrF & - & 17.91 & 58.84 & 53.95 \\
    \end{tabular}
    \caption{Diversity metrics for the Turkish BT datasets (original language: English) used to train the Tr$\rightarrow$En models. Inter-sentence metrics are calculated on a sample of 30k triplets. `M' = million.}
    \label{tab:tren-diversity}
\end{table}

\begin{table}[t!]
\small
    \centering
    \begin{tabular}{rccccc}
        Dataset & \textit{base-big} & \textit{beam} & \textit{sampling} & \textit{nucleus} \\
        \midrule
        Sent. len. & 15.65 & 14.79 & 14.92 & 14.73 \\
        Word len. & 6.54 & 6.91 & 7.33 & 7.15 \\
        Vocab.& 1.3M & 0.82M & 11M & 5.6M \\
        \midrule
        i-BLEU & - & 30.89 & 86.41 & 79.67 \\
        i-chrF & - & 16.09 & 66.06 & 57.83 \\
    \end{tabular}
    \caption{Diversity metrics for the Icelandic BT datasets (original language: 
    English) used to train the Is$\rightarrow$En models. Inter-sentence metrics are calculated on a sample of 30k triplets. `M' = million.}
    \label{tab:isen-diversity}
\end{table}

\subsection{Diversity metrics}

We give the diversity metrics for each language pair and each generated dataset in \Cref{tab:tren-diversity,tab:entr-diversity,tab:isen-diversity,tab:enis-diversity}.\footnote{We omit \textit{base} for reasons of space and because its different length to the other datasets makes comparison difficult (3 million lines compared to 9 million for the others).} Sentence and word lengths are comparable across the same language for all generation methods, suggesting that each method is generating tokens from roughly the right language distribution. However, the vocabulary size is much larger for \textit{nucleus} compared to \textit{base} or \textit{beam}, and \textit{sampling} is around twice that of \textit{nucleus}. Examining the data, we find many neologisms (that is, `words' which do not appear in the training data) for \textit{nucleus} and more still for \textit{sampling}. We note that the \textit{syntax-groups} dataset has a much smaller vocabulary again; this is what we would hope if the generation method is producing syntactic rather than lexical diversity as required. We give representative examples of generated triples in \Cref{sec:triple_ex}, along with some explanation of how the phenomena they demonstrate fit into the general trend of the dataset. 

\paragraph{Effect on performance} With respect to the inter-sentence diversity metrics (i-BLEU, i-chrF, and tree kernel scores), we see that the \textit{sampling} dataset has the highest diversity scores, followed by \textit{nucleus}, then \textit{syntax}, then \textit{beam}. Taken together with the performance scores and the summary statistics, this suggests that \ac{NMT} data benefits from a high level of diversity, but not so high that the two halves of the parallel data no longer have the same meaning (as shown by the very high vocabulary size for \textit{sampling}).

\begin{table}[t!]
\small
    \centering
    \begin{tabular}{rcccccc}
        Dataset & \textit{base+} & \textit{beam} & \textit{sampl.} & \textit{nucleus} & \textit{syntax} \\
        \midrule
        Sent. len. & 16.98 & 17.03 & 17.85 & 17.54 & 17.28 \\
        Word len.  & 6.05 & 6.04 & 6.25 & 6.11 & 6.06 \\
        Vocab & 0.89M & 0.54M & 4.9M & 2.5M &  0.64M \\
        \midrule
        i-BLEU & - & 30.74 & 83.52 & 78.92 & 42.26 \\
        i-chrF & - & 16.28 & 57.16 & 51.82 & 23.32 \\
        Kernel & - & 72.20 & 97.33 & 95.91 & 83.43 \\
    \end{tabular}
    \caption{Diversity metrics for the English BT datasets (original language: Turkish) used to train the En$\rightarrow$Tr models. Inter-sentence metrics are calculated on a sample of 30k triplets. `M' = million.}
    \label{tab:entr-diversity}
\end{table}

\begin{table}[t!]
\small
    \centering
    \begin{tabular}{rcccccc}
        Dataset & \textit{base+} & \textit{beam} & \textit{sampl.} & \textit{nucleus} & \textit{syntax} \\
        \midrule
        Sent. len. & 20.45 & 22.75 & 21.34 & 21.13 & 18.29 \\
        Word len. & 5.83 & 5.83 & 6.33 & 6.08 & 5.89 \\
        Vocab. & 0.66M & 0.41M & 12M & 5.6M & 0.49M \\
        \midrule
        i-BLEU  & - & 22.75 & 92.31 & 88.86 & 77.17 \\
        i-chrF  & - & 11.95 & 72.20 & 67.16 & 56.90 \\
        Kernel & - & 65.72 & 99.35 & 98.74 & 99.40 \\
    \end{tabular}
    \caption{Diversity metrics for the English BT datasets (original language: 
    Icelandic) used to train the En$\rightarrow$Is models. Inter-sentence metrics are calculated on a sample of 30k triplets. `M' = million.}
    \label{tab:enis-diversity}
\end{table}

\paragraph{Metric correlation} There is a high correlation between i-BLEU, i-chrF, and tree kernel score for the \textit{beam}, \textit{sampling}, and \textit{nucleus} datasets. This is not entirely unexpected: it is likely to be difficult if not impossible to disentangle lexical and syntactic diversity, since changing sentence structure would also affect the word choice and vice versa. 

This correlation is much weaker for the \textit{syntax-groups} dataset: whilst the tree-kernel scores are comparable to the \textit{sampling} and \textit{nucleus} datasets, there is a much smaller increase in the other (lexical) diversity scores. This suggests that this generation method encourages relatively more syntactic variation than lexical compared to the other diverse generation method, as was its original aim (see paragraph on syntax-group fine-tuning in \cref{sec:generating}). The fact that the final model trained on this \ac{BT} dataset has lower performance compared to other forms of diversity suggests that lexical diversity is more important than syntactic diversity when undertaking data augmentation. We leave it to future work to investigate this hypothesis further.

\subsection{Data augmentation versus more monolingual data}
The right-most cross in each quadrant of \Cref{fig:perf_bleu,fig:perf_comet} gives the performance of \textit{base-big}, the dataset where we simply add six million more lines of new data rather than carrying out data augmentation. Interestingly, pure and nucleus sampling both often outperform \textit{base-big}. This may be because the model over-fits to too much back-translated data, whereas having multiple sufficiently-diverse pseudo-source sentences for each target sentence has a regularising effect on the model. 

\begin{figure}[htbp]
    \centering
    \includegraphics[width=\linewidth]{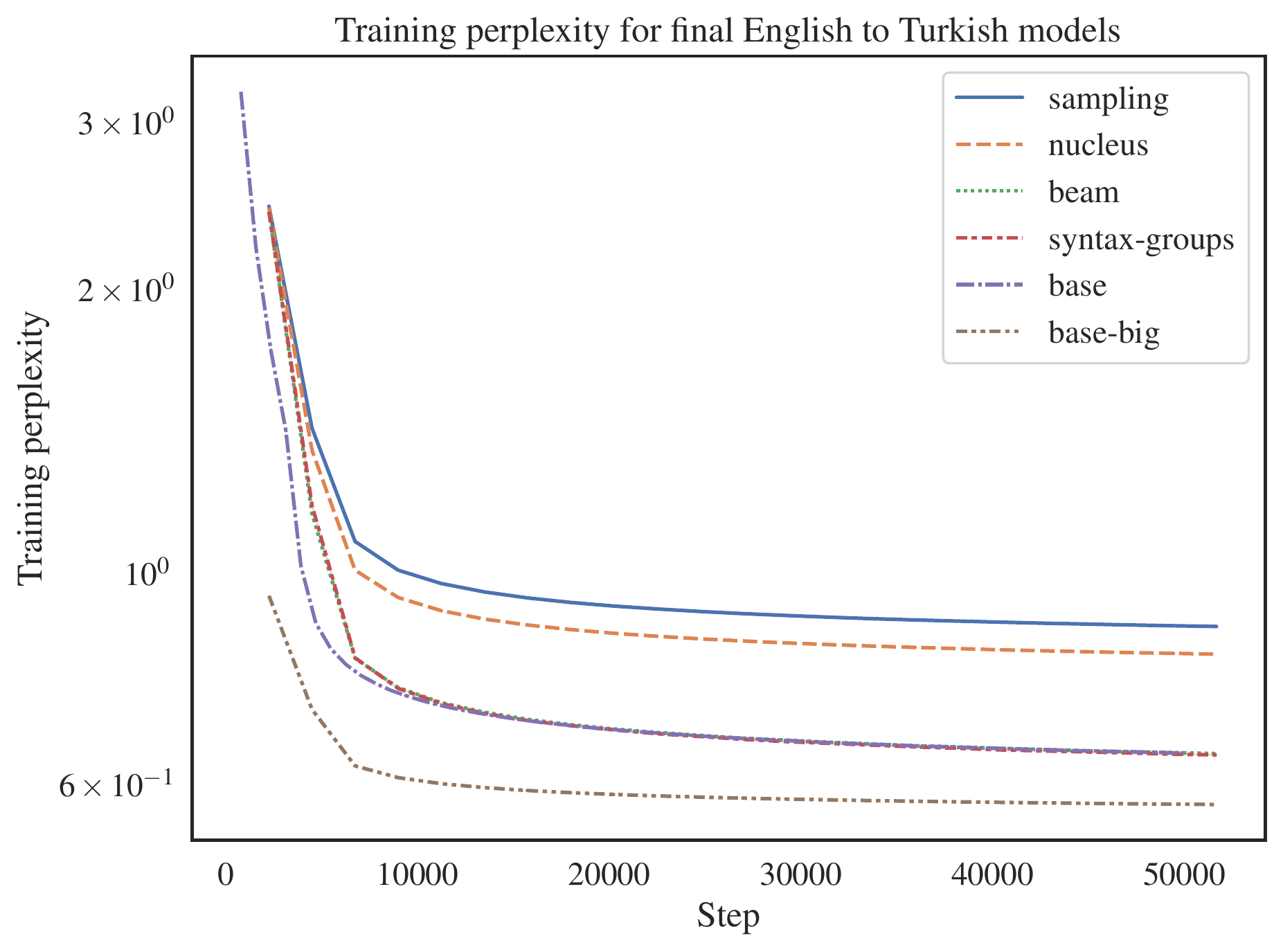}
    \caption{Mean training perplexity for the first 50 thousand steps of training for final English$\rightarrow$Turkish models. The model has highest training perplexity on the \textit{sampling} then \textit{nucleus} datasets. The lowest training perplexity is on the \textit{beam} and \textit{base-big} datasets.}
    \label{fig:entr-train-curve}
\end{figure}

To further support this hypothesis, \Cref{fig:entr-train-curve} gives training perplexity for the first 50,000 steps of training for the final Icelandic$\rightarrow$English models, which are representative of the results for the other language pairs. We see that the \textit{base-big} dataset has the lowest training perplexity at each step, suggesting this data is easier to model. Conversely, the model has highest training perplexity on the \textit{sampling} and \textit{nucleus} datasets, suggesting generating the data this way has a regularising effect.

\subsection{Translationese effect}

Several studies have found that back-translated text is easier to translate than forward-translated text, and so inflates intrinsic metrics like BLEU \citep{edunov-etal-2020-evaluation, graham-etal-2020-assessing, roberts-etal-2020-decoding}. To use a concrete example, the WMT test sets for English to Turkish are made up of half native English translated into Turkish, and half native Turkish translated into English. We want models that perform well when translating \textit{from} native text (in this example: the native English side), as this is the usual direction of translation. However, half the test set is made up of translations on the source side. The translationese effect means that the model will usually get higher scores on this half of the test set, potentially inflating the score. Consequently, the intrinsic metrics could suggest choosing a model that does not actually perform well on the desired task (translating \textit{from} native text).

\begin{figure*}[htbp]
    \centering
    \includegraphics[width=\textwidth]{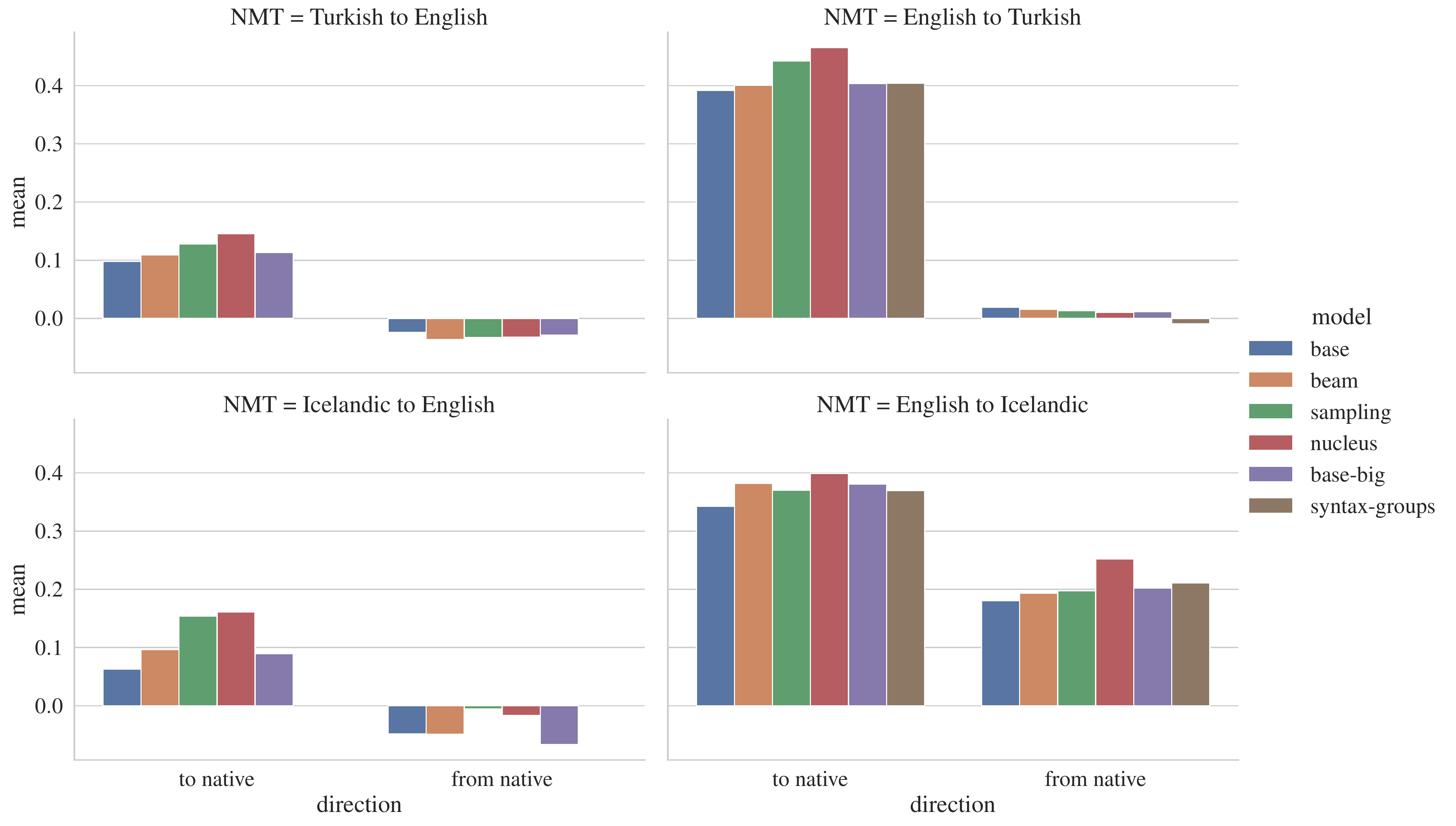}
    \caption{The mean percentage change in BLEU score for each model on the test set(s) over the parallel-only models, separated by language direction. The left-hand side (\textit{to native}) has translated text on the source side and native text on the target side of the test set (back translation). The right-hand side (\textit{from native}) has native text on the source side and translated text on the target side of the test set.}
    \label{fig:bleu_trans}
\end{figure*}

We investigate this effect in our own work by examining the mean BLEU scores for each model on each half of the test sets, giving the results in \Cref{fig:bleu_trans}. Each bar indicates the mean percentage change in BLEU scores over the parallel-only baseline model for the models trained on the different \ac{BT} datasets, so a larger bar means a better performing model. The left-hand bars in each quadrant show the performance of each model on the back-translated half of the test set (\textit{to native}) and the right-hand bars give the performance of each model on the forward-translated half of the test set (\textit{from native}).

We see a significant translationese effect for all models, as the percentage change in scores over the baseline are much higher when the models translate already translated text (the left-hand side bars are higher than the right-hand ones). However, it appears that the \textit{nucleus} dataset is less affected by the translationese effect than the other datasets, since it shows less of a decline in performance when translating native text. This may be due to a similar regularising effect as discussed previously, as it is more difficult for the model to overfit to \ac{BT} data when it is generated with nucleus sampling. A direction for future  research is how to obtain the benefits of using monolingual data (as \ac{BT} does) without exacerbating the translationese effect.

\section{Related work}
\paragraph{Improving back translation}
The original paper introducing \ac{BT} by \citet{sennrich-etal-2016-improving} found that using a higher-quality \ac{NMT} system for \ac{BT} led to higher BLEU scores in the final trained system. This finding was corroborated by \citet{burlot-yvon-2018-using}, and following work has investigated further ways to improve \ac{NMT}. These include iterative \ac{BT} \citep{hoang-etal-2018-iterative}, targeting difficult words \citep{fadaee-monz-2018-back}, and tagged \ac{BT} \citep{caswell-etal-2019-tagged}. Section 3.2.1 of \citet{haddow-etal-2021-survey} presents a comprehensive survey of \ac{BT} and its variants as applied to low-resource \ac{NMT}.

\paragraph{Diversity in machine translation}
Most of the work on the lack of diversity in machine-translated text are in the context of automatic evaluation \citep{edunov-etal-2020-evaluation, graham-etal-2020-assessing, roberts-etal-2020-decoding}. As for diversity in \ac{BT} specifically, \citet{edunov-etal-2018-understanding} argue that \ac{MAP} prediction, as is typically used to generate \ac{BT} through beam search, leads to overly-regular synthetic source sentences which do not cover the true data distribution. They propose instead generating \ac{BT} with sampling or noised beam outputs, and find model performance increases for all but the lowest resource scenarios. Alternatively, \citet{soto-etal-2020-selecting} generate diverse \ac{BT} by training multiple machine-translation systems with varying architectures. 

\paragraph{Generating diversity}
Increasing diversity in \ac{BT} is part of the broader field of diverse generation, by which we mean methods to vary the surface form of a production whilst keeping the meaning as similar as possible. Aside from generating diverse translations \citep{gimpel-etal-2013-systematic, he-etal-2018-sequence, shen-etal-2019-mixture, nguyen2020data, li-etal-2021-mixup-decoding}, it is also used in question answering systems \citep{sultan-etal-2020-importance}, visually-grounded generation \citep{vijaykumar-etal-2018-diverse}, conversation models \cite{li-etal-2016-diversity}, and particularly paraphrasing \citep{mallinson-etal-2017-paraphrasing, wieting-gimpel-2018-paranmt, hu2019parabank, thompson-post-2020-paraphrase,goyal-durrett-2020-neural,krishna-etal-2020-reformulating}. Some recent work such as \citet{iyyer-etal-2018-adversarial}, \citet{huang-chang-2021-generating}, and \citet{hosking-lapata-2021-factorising} explicitly model the meaning and the form of the input separately. In this way, they aim to vary the syntax of the output whilst preserving the semantics so as to generate more diverse paraphrases. Unfortunately, these methods are difficult to apply to a low-resource scenario as they require external resources (e.g. accurate syntactic parsers, large-scale paraphrase data) which are not available for most of the world's languages.

\section{Conclusion}

In this paper, we introduced a two-part framework for understanding diversity in \ac{NMT} data, splitting it into \textit{lexical diversity} and \textit{syntactic diversity}. Our empirical analysis suggests that whilst high amounts of both types of diversity are important in training data, lexical diversity may be more beneficial than syntactic. In addition, achieving high diversity in \ac{BT} should not be at the expense of adequacy. We find that generating \ac{BT} with nucleus sampling results in the highest final \ac{NMT} model performance for our systems. Future work could investigate further the affect of high lexical diversity on \ac{BT} independent of syntactic diversity.

\section*{Acknowledgements}

This work was supported in part by the UKRI Centre for Doctoral Training in Natural Language Processing, funded by the UKRI (grant EP/S022481/1) and the University of Edinburgh, School of Informatics and School of Philosophy, Psychology \& Language Sciences. It was also supported by funding from the European Union’s Horizon 2020 research and innovation programme under grant agreement No 825299 (GoURMET) and funding from the UK Engineering and Physical Sciences Research Council (EPSRC) fellowship grant EP/S001271/1 (MTStretch). 

The experiments in this paper were performed using resources provided by the Cambridge Service for Data Driven Discovery (CSD3) operated by the University of Cambridge Research Computing Service (www.csd3.cam.ac.uk), provided by Dell EMC and Intel using Tier-2 funding from the Engineering and Physical Sciences Research Council (capital grant EP/P020259/1), and DiRAC funding from the Science and Technology Facilities Council (www.dirac.ac.uk). 

Finally, the authors would like to thank our anonymous reviewers for their time and helpful comments, and we give special thanks to Henry Conklin and Bailin Wang for their help with tree kernels and many useful discussions.

\bibliography{anthology,custom}
\bibliographystyle{acl_natbib}
\clearpage

\appendix
\section{Appendix}  % needs to be one page
\label{sec:appendix}

\subsection{Representative examples from back-translated datasets (translated from Icelandic)}
\label{sec:triple_ex}

\paragraph{Original}
\textit{Þjóðverjar hafa tekið forræðið og stefnt er að stofnun stórríkis.}

\subsubsection*{Beam}
\begin{itemize}
    \item The Germans have taken custody and are aimed at \ul{the establishment} of a large state.
    \item The Germans have taken custody and are aimed at \ul{the creation} of a large state.
    \item The Germans have taken custody and are aimed at \ul{establishing} a large state.
\end{itemize}

\textit{Comment:} Only one or two words differ between sentences (underlined).

\subsubsection*{Sampling}
\begin{itemize}
    \item The \ul{Germz} \ul{governmentregluru} has committed suicide, intending to organise a major state.
    \item The \ul{Germano} had ensured that British commanders in France would be aides of \ul{the\ae{}rd} rapidly.
    \item And the need to defend and establish \ul{theseUCtions} are all organized \ul{intomissions} from \ul{Iraq\'e{}ttihe}.
\end{itemize}

\textit{Comment:} Sentences show large variation  in structure and vocabulary, but they contain many non-dictionary words (underlined) and adequacy is low. 

\subsubsection*{Nucleus}
\begin{itemize}
    \item Germany has taken custody and aimed to establish a large country.
    \item The German government initiated a \ul{group operation}, to establish capital city.
    \item The Germany has managed to make an example of their full \ul{widowed demands}.
\end{itemize}

\textit{Comment:} There is a moderate amount of variation between sentences in terms of syntax and vocabulary, but no non-dictionary words. Some phrases lack adequacy (underlined).

\subsubsection*{Syntax-groups}
\begin{itemize}
    \item The Germans have taken custody and are aimed at the establishment of a large state.
    \item The \ul{Icelandic Institute of Natural History}
    \item \ul{As a result,} the Germans have taken control of the country and are aimed at establishing a large state.
\end{itemize}

\textit{Comment:} The second and third sentences contain hallucinations, presumably in order to generate according to the syntactic templates (underlined).

\subsection{Model architecture and infrastructure}
\label{sec:model_desc}

All \ac{NMT} models in this paper are transformer models \citep{vaswani-etal-2017-attention}. We conducted a hyper-parameter search for each language pair, training English$\leftrightarrow$Turkish and English$\leftrightarrow$Icelandic \ac{NMT} models and using the BLEU score as the optimisation metric. We give the settings which differ to \textit{transformer-base} in \Cref{tab:hyp}. We use the same hyper-parameter settings for all models trained for the same language pair. 

We use the Fairseq toolkit to train all our \ac{NMT} models \citep{ott-etal-2019-fairseq}. We train on four NVIDIA A100-SXM-80GB GPUs and use CUDA 11.1 plus a Python 3.8 Conda environment provided in the Github repo. We generate on one GPU, since to our knowledge the Fairseq toolkit does not support multi-GPU decoding. We use Weights and Biases for experiment tracking \citep{wandb}.

\begin{table}[htbp]
    \small
    \centering
    \begin{tabular}{rcc}
         & tr--en & is--en \\
         \hline
         Dropout & 0.6 & 0.3 \\
         Activation dropout & 0.1 & 0 \\
         Attention dropout & \multicolumn{2}{c}{0.1} \\
         Learning rate & \multicolumn{2}{c}{0.001} \\
         L.R. scheduler & \multicolumn{2}{c}{Inv. square root}\\
         Optimiser & \multicolumn{2}{c}{Adam} \\
         Optimiser parameters & \multicolumn{2}{c}{0.9, 0.98} \\
         Label smoothing & \multicolumn{2}{c}{0.1} \\
         Shared embeddings & \multicolumn{2}{c}{all} \\
         Batch size & \multicolumn{2}{c}{64} \\
         Update frequency & \multicolumn{2}{c}{16} \\
         Patience & \multicolumn{2}{c}{15} \\
    \end{tabular}
    \caption{Hyper-parameter settings for NMT transformer models trained for each language pair. All other settings are the default for \textit{transformer-base}.}
    \label{tab:hyp}
\end{table}

\end{document}